\documentclass[letterpaper, 10 pt, conference]{ieeeconf}  
\usepackage{amsmath,amsfonts}
\usepackage{amssymb}
\usepackage{xpatch}
\usepackage{algorithm}
\usepackage{algpseudocode}
\usepackage{array}
\usepackage[caption=false,font=normalsize,labelfont=sf,textfont=sf]{subfig}
\usepackage{textcomp}
\usepackage{stfloats}
\usepackage{url}
\usepackage{verbatim}
\usepackage{graphicx}
\usepackage{tabularx}
\usepackage{cite}
\usepackage{url}
\usepackage[hidelinks=true]{hyperref}
\usepackage[T1]{fontenc}
\usepackage{textcomp}
\usepackage{gensymb}
\usepackage{color}
\usepackage{csquotes}
\algrenewcommand\textproc{}

\IEEEoverridecommandlockouts                              

\title{Layered Cost-Map-Based Traffic Management for Multiple Automated Mobile Robots via a Data Distribution Service}

\author{Seungwoo Jeong$^{1}$,
	Taekwon Ga$^{1}$,
	Inhwan Jeong$^{2}$,
	Jongkyu Oh$^{2}$, 
	and Jongeun Choi$^{1}$,~\IEEEmembership{Member,~IEEE}
	\thanks{This study was supported by Hyundai Robotics. All the robots were tested and demonstrated in the Hyundai Robotics laboratory, and environmental settings were configured in the same laboratory.			
	$^{1}$Seungwoo Jeong, Taekwon Ga, and Jongeun Choi are with the School of Mechanical Engineering, Yonsei University, Seoul 03722, Republic of Korea (e-mail: slsw@yonsei.ac.kr; taek111@yonsei.ac.kr; jongeunchoi@yonsei.ac.kr). $^{2}$Inhwan Jeong and Jongkyu Oh are with Hyundai Robotics, Yong-in 16891, Republic of Korea (e-mail: 123inani@hyundai-robotics.com; jkoh@hyundai-robotics.com). The corresponding author is Jongeun Choi.}
}

\begin{document}

\maketitle
\thispagestyle{empty}
\pagestyle{empty}

\begin{abstract}
This letter proposes traffic management for multiple automated mobile robots (AMRs) based on a layered cost map. Multiple AMRs communicate via a data distribution service (DDS), which is shared by topics in the same DDS domain. The cost of each layer is manipulated by topics. The traffic management server in the domain sends or receives topics to each of AMRs. Using the layered cost map, the new concept of prohibition filter, lane filter, fleet layer, and region filter are proposed and implemented. The prohibition filter can help a user set an area that would prohibit an AMR from trespassing. The lane filter can help set one-way directions based on an angle image. The fleet layer can help AMRs share their locations via the traffic management server. The region filter requests for or receives an exclusive area, which can be occupied by only one AMR, from the traffic management server. All the layers are experimentally validated with real-world AMRs. Each area can be configured with user-defined images or text-based parameter files.

\end{abstract}

\section{INTRODUCTION}
Traffic management is essential for an automated mobile robot (AMR) to effectively perform tasks. A typical example of a fixed task is that an AMR starts from a point (the origin) and performs a set of tasks at the destination points. 
While operating multiple AMRs, a set of traffic rules can be useful in opertating multiple AMRs in a factory or warehouse. For example, when human workers and AMRs coexist in a factory or warehouse, it would be more efficient for the AMRs to perform tasks based on right-hand traffic if the AMRs pass through the corridor with humans.
In this case, if the corridor is divided into two passages and a one-way passage is introduced in one of the passages, the AMR can be forced to take the right-hand route in that passage. 
It is common for multiple AMRs to get stuck in a deadlock. 
Unlike an automated guided vehicle (AGV), which follows a set path which is similar to a train that follows the rail route, an AMR does not possess a specific traffic management algorithm as there are no fixed paths. Thus, if two or more AMRs enter a narrow passage, where only one AMR can pass, at the same time, they fall into a deadlock state, which needs to be manually removed by a human traffic manager. In addition, when one AMR performs a task in a specific area and another AMR performs a task that interferes with the working AMR, the efficiency of the working AMR is reduced. If the AMR that arrives first has a high task priority, an extreme case may occur; e.g., the next AMR would not complete the task within the set time and conduct the mission normally. To cope with the traffic management of multiple AMRs, the layered cost map based on the data distribution service (DDS) is proposed; it can be intuitively and efficiently controlled by the user or AMR.

The contributions of this letter are as follows: A novel traffic management approach of multiple AMRs is proposed and implemented by using a layered  cost map. The information of AMRs is shared as a form of topics using a DDS. In contrast to the existing method, which requires separate network settings based on a TCP/IP and a sensor network, a new AMR can easily participate in collaboration with other AMRs  using topics if the AMR is  in the same DDS domain. Our approach can allows the user to set  a prohibited area or one-way area. The conventional AMR collision avoidance method can be performed using cameras or lidar sensors. However, the method must be accompanied by trajectory prediction for the uncertain collision avoidance strategy. The fleet layer is used to effectively avoid collision  among AMRs   sharing information with the DDS. 
To effectively operate multiple AMRs, the region filter is used to reserve an area for only one AMR, whereas the remaining AMRs can wait outside of the area. Requests and returns for the area are recorded in the global data space using DDS topics  so that all the AMRs can share the region. 
Using the region filter, the function of allowing only one AMR to pass through a narrow passage has been effectively implemented. 
The region filter is set by the convex region as an array. Experiments with real AMRs have been performed to validate the proposed algorithms  of the prohibition filter,   lane filter, fleet layer, and region filter.

\section{RELATED WORKS}
Over the past years, several studies have focused on the traffic management of AGVs. To resolve collision and deadlock, the transport road network is applied to divide non-overlapping zones \cite{ZAJAC202180}. The path planning for quay and rail-mounted gantry cranes with AGVs has been integrated with the mixed-integer programming model based on path optimization, integrated scheduling, conflicts and deadlock \cite{ZHONG2020106371}. For highly scalable management, AGVs autonomously execute pick-up and delivery operations based on a fully decentralized control algorithm \cite{DRAGANJAC2020101915, FANTI201886}. A traffic management
method with a virtual network map, considering time constraints, has been added to the planning process \cite{9241435}. A hierarchical layer control architecture has been adopted for the traffic management of AGVs \cite{9560828}. Using the robotics operating system (ROS), the experiment has been conducted to verify a solution to the scheduling deadlock problem using the rotational anti-deadlock algorithm \cite{9659172}. A self-adaptive traffic management model with behavior trees and
reinforcement learning (RL) has been proposed to make optimal
decisions to cope with diverse, dynamic, and complex situations \cite{9355019}.

The cost map is often used for obstacle avoidance in robot navigation. 
The obstacle depicts a high-cost value in the cost map based on its location and speed. 
The cost map generally helps in navigation by separating dynamic obstacles from static ones. A typical dynamic obstacle can be defined for a person who walks across the robot\textquotesingle s path \cite{fang2020human, 9636613, 9561522, 7140063, 8460978, 7324184, 7745154, 9515475}. 
Predicting patterns of pedestrians  and reflecting them in the cost map can help further reduce the deadlocks between the robots and humans. 
Several studies have been presented, in which the human behavior is predicted and reflected in the cost map \cite{8036225}. The social robot placement not to interfere with human workers   could be  modeled and reflected in the cost map \cite{9340892}. Few studies have accurately predicted the human trajectory using deep inverse RL  \cite{7487452, lim2020prediction}. 
Moreover, the cost map has been used to coordinate multiple robots \cite{8460978}. 
Ellis et al. proposed risk averse bayesian reward learning for autonomous navigation from human demonstration  \cite{9635835}.

Sivaprakasam et al. have  investigated a scenario in which the robot navigates to an optimized path by reflecting the terrain in the cost map \cite{9561881}. 
In an off-road environment, the robot calculates the cost of the path so that it  can successfully pass avoiding obstacles with a high probability. Ugur and Bebek \cite{9551617} presented the motion planning of the exploration rover based on a real-time cost map using a depth field and color image data to  pass through rugged terrain. Similarly,  a study \cite{9340912} proposed an approach that divides  terrain   into paths, and   assigns each of paths  with a cost for path planning  in a mars-like environment. In an indoor environment, a robot created a 3D-octomap using wheel odometry, 2D laser, and RGB-D camera to pass through the terrain  so that the robot moves up and down the slope, avoiding the stairs, arriving  at a set destination \cite{8202145}. Moreover, Regier et al. presented an approach that  a robot analyzed the density of regions (rather than the terrain) in order to  pass  through a region that was not dense \cite{7759234}.

\section{METHODS}
\subsection{Data Distribution Service}

\begin{figure}[]
	\begin{center}
		\includegraphics[width=8.5cm]{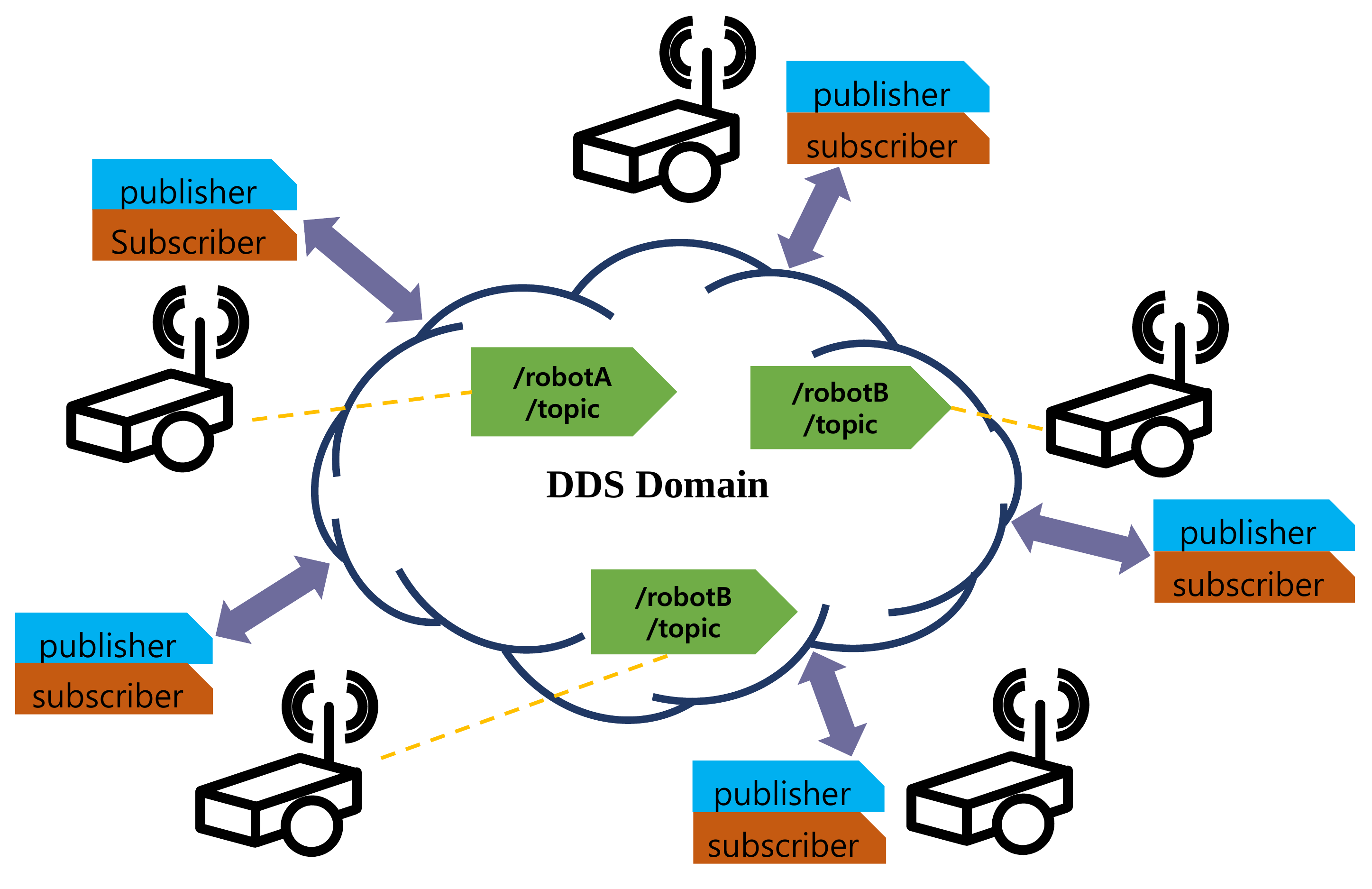}
		\caption{Data sharing in the global data space in the DDS domain}
		\label{ddsconcept}
		\vspace{-0.6cm}
	\end{center}
\end{figure}

In a distributed system environment, a DDS \cite{1203555} is data-centric, and all the participating AMRs in the DDS domain are allowed to be data publishers and data subscribers, as shown in Fig. \ref{ddsconcept}. Other participating AMRs can read the data written by any   of the AMRs in the same DDS domain. In the DDS, the standard data type is described by a topic name and value pair $topic=<name, value>$. All the data are virtually stored in a global data space. The global data space may be in a distributed set of hardware units or the local memory of individual robots. 
A topic is transmitted in a multicast manner to share data reliably. To identify which topics are involved, an identifier, e.g., a namespace, is prefixed to the topic name in the following style: $/namespace/topic$. A namespace corresponds to the name of an AMR. 

In this letter, a single centralized traffic management server is applied to set the common configurations of AMRs. To send maps, the DDS enables the traffic management  server to simultaneously send static or user-defined maps as topics to multiple AMRs. In contrast to  AMRs that pre-download maps, the participating AMRs in the DDS can share maps with the traffic management. Moreover, AMRs can communicate with each other to share the information of their positions or occupied regions. All the AMRs send their position information to the single traffic management server with the DDS topic in real time to avoid collision. To control the region occupation, a ticket that includes the robot and region information is shared bidirectionally between the AMRs and the traffic management server.

\begin{figure*}[t]
	\begin{center}
		\includegraphics[width=17.5cm]{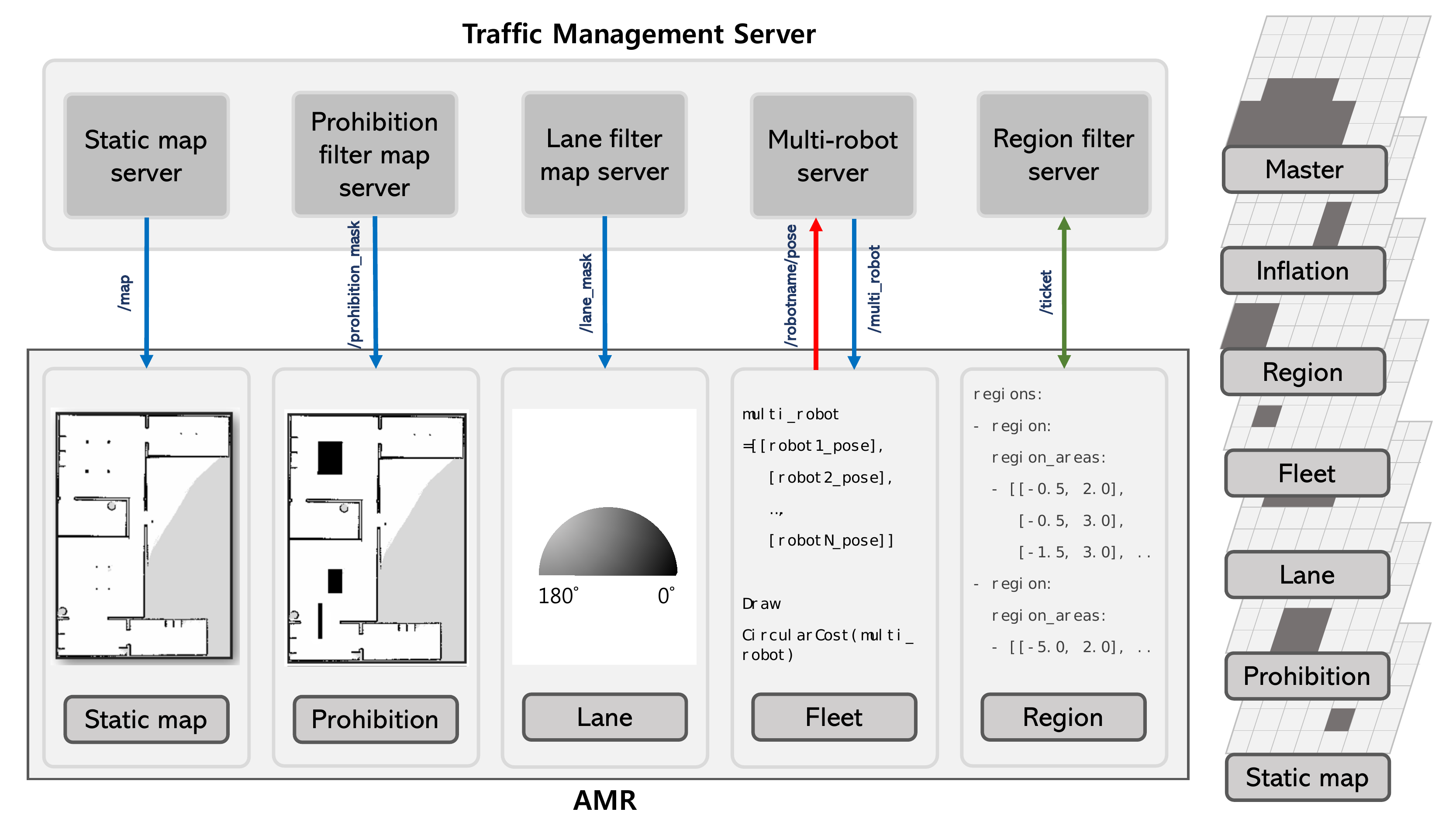}
		\vspace{-0.3cm}
		\caption{Layered cost map: the prohibition filter, lane filter, fleet layer, and region filter}
		\label{layeredcostmap}	
		\vspace{-0.5cm}
	\end{center}
\end{figure*}
\subsection{Layered Cost Map}
The cost map   demonstrates a discrete space instead of a continuous space (Fig. \ref{layeredcostmap}). Our  AMRs calculate  the cost of each cell  so that  AMRs move toward the path with the lowest cost. AMR planning is categorized into global planning that  does not consider dynamic obstacles, and local planning that  considers dynamic obstacles. 
The algorithm using the cost map, known as the dynamic window approach (DWA)  for global planning and local planning \cite{580977} is adopted in this letter. 
This algorithm is positioned as the basic planning algorithm of ROS1 and ROS2, considering its strength that the computational cost is extremely low. The cost map expresses   objects or spaces that   AMRs should avoid with regard to the cost. The layered cost map is expressed using OR operation by stacking multiple cost maps for obstacles and spaces \cite{6942636}. 
Similar to ROS1, ROS2 provides the layered cost map as a plugin library. Once a plugin library is created, a plugin is referenced for global and local planners. The most basic layer is the static map, which receives a map drawn from  simultaneous localization and mapping (SLAM). It helps   AMRs avoid walls and static obstacles. 
Although static and dynamic obstacles are reflected as a form of  cost values,  AMRs may  approach very closely  to those obstacles. 
To increase the separation distance for navigation planning, it is necessary to express an area around the obstacle with a lower cost value compared with the obstacle cost to provide buffer areas  before collision. 
This area is represented by the inflation layer. 
One of our goals   is   to help multiple AMRs proceed to their destinations without any collision; thus, the fleet layer has been designed to share the locations among the AMRs. 
In addition, the master cost map grid stacks separate layers, including  the static map layer, prohibition filter, lane filter, fleet layer, region filter, and inflation layer. The prohibition filter, lane filter fleet layer, and region filter are stacked for no-trespassing, one-way, and collision avoidance and exclusive areas, respectively. 

\subsubsection{\textbf{Prohibition filter}} 
Let us consider an area where    no obstacles  exist  but  high risk locations  exist for AMRs.
Such an area can be  designated as a prohibited area  by user-defined image to prevent  AMRs from passing through it. As   AMRs  are prevented from trespassing this area, the cost map layer is named   the prohibition filter. 
The cost exhibits a maximum level of 254. If necessary, it can be reduced to a non-maximum level  so that the AMR can flexibly trespass the area when it needs to avoid dynamic obstacles. We designated a prohibition filter by adding the filter part as a suffix because a separate map was required. To contrast this map against  a static map, we call the map to the prohibition filter mask map. Similar to the static map, the mask map is transmitted from a separate map server and delivered to the topic---\texttt{/prohibition\_mask}. 
Then, the prohibition filter is created from the mask map and is overlapped with the existing lower cost map layers using OR operation. 
As shown in Fig. \ref{layeredcostmap}, the mask map is created with the static map by adding the prohibited area, which is marked in black.

\subsubsection{\textbf{Lane filter}}
\begin{algorithm}[t]
	\caption{Lane filter}\label{alg:lanefilter}
	\begin{algorithmic}
	\State {\textbf{Input} : $\angle$ robot; the yaw angle of the robot}
	\State {\phantom{Input : } $\angle$ lane; pixel color value in range 0 to 35999}
	\State \textbf{Output} : Cost
		\Function{GetCost}{$\angle$ robot $, \angle$ lane}
		\State Wait the response of the action server
		\If {$0.4 \leq \cos \left(\angle_{\text {robot }}-\angle_{\text {lane }}\right) \leq 1.0$}
		\State {Cost = 0}
		\ElsIf {$-1.0 \leq \cos \left(\angle_{\text {robot }}-\angle_{\text {lane }}\right) \leq-0.4$}
		\State {Cost = 254}
		\Else \State {Cost = 128}
		\EndIf
		\State \Return {Cost}
		\EndFunction
	\end{algorithmic}
\end{algorithm}

 AMRs can effectively pass through a long and narrow passageway with people or other AMRs simultaneously if AMRs are regulated to right-hand traffic. The lane filter is in charge of enforcing such a  traffic rule. The lane filter uses a map server that requires a separate mask map file to send maps as topics---\texttt{/lane\_mask}. However, one difference is that it receives a 16-bit map file to set directions. 
The lane filter map server is designed to receive the 16-bit map file to set the 16-bit direction cost map. 
Similar to the mask map of the prohibition filter, the 16-bit mask map is converted to a cost map. The cost can be dynamically adjusted according to the yaw angle of the AMR, as described in Algorithm \ref{alg:lanefilter}.

\subsubsection{\textbf{Fleet layer}}
In this study, the layer that adjusts the surrounding cost map based on the locations of AMRs to prevent collision with other AMRs by sharing their locations is named the fleet layer. We designated a fleet layer by adding the layer part as a suffix because a separate map or parameters was not required as compared to the prohibition filter.
$N$ AMRs share their locations as the coordinate transforms topic--- \texttt{/multi\_robot}. 
Each AMR then marks the position  of the other AMR  in  the  cost map with  a high cost value, i.e., a 2D circle-shaped cost plus the inflated cost, which is bell-shaped in 3D. As a result, each robot can avoid collision.

The collision avoidance is based on the basic DWA. 
This algorithm has been widely used for a single AMR owing to its simplicity and low computational cost. DWA is a cost-map-based algorithm, which is effective for static obstacle avoidance. Dynamic obstacles in front of an AMR are recognized and can be  reflected in the cost map.
To reflect multiple AMRs with the high cost values, it is necessary to collect the pose  information of each AMR as a topic---\texttt{/robotname/pose} in the multi-robot server. 

\subsubsection{\textbf{Region filter}}
\begin{algorithm}[t]
	\caption{Region filter}\label{alg:regionfilter}
	\begin{algorithmic}
		\State {\texttt{ticket} : robot $ID$, region $ID$, return value}
		\\
		\State {\textbf{AMR side :}}
			\State {Upon entering the region :}
			\If {Is AMR in the inflation region}
			\State {response=reserveRegion(robot $ID$, region $ID$)}
				\If {response==Success}
					\State{Perform the given task}
				\EndIf
			\EndIf
			
			\State {Upon leaving the region :}
			\If {Is AMR leaving out of the inflation region}
			\State {response=releaseRegion(robot $ID$, region $ID$)}
				\If {response==Success}
				\State{Send the release to the region filter server}
				\EndIf
			\EndIf
		\\
		\State {\textbf{Region filter server side :}}
		\Function{reserveRegion}{robot $ID$, region $ID$}
			\State{region\_info=SearchAMR(robot $ID$, region $ID$)}
			\If{region\_info==empty}
				\State{AllocateAMR(robot $ID$, region $ID$)}
				\State \Return {Success}
			\EndIf
		\EndFunction
		
		\Function{releaseRegion}{robot $ID$, region $ID$}
			\State{region\_info=SearchAMR(robot $ID$, region $ID$)}
			\If{region\_info!=empty}
			\State{DeallocateAMR(robot $ID$, region $ID$)}
			\State \Return {Success}
			\EndIf
		\EndFunction
	\end{algorithmic}

\end{algorithm}

In a multi-AMR system,  there could be limits on the number of AMRs at a certain area.
For example, if two AMRs enter a narrow road, which is difficult to pass simultaneously, they may enter a deadlock situation. 
To cope with this situation, the region filter can be  designed to prevent deadlock when  multiple AMRs enter the same area simultaneously. The region is represented by the convex hull. The convex hull region is configured with  a set of vertices, which defines an array type in the YAML file. 
The configuration of the region filter is obtained using the text-based parameter file. The region list includes x and y coordinate vectors with [double, double] type. 
When an AMR enters a pre-designated area by setting an exclusive area in the cost map, it changes the free space to an occupied space, which prohibits other AMRs; thus, we define the layer as the region filter. In the region filter, when one AMR enters the exclusive area, the information must be reflected in the cost map of other AMRs; thus, location sharing among AMRs is required.

The region filter communicates with the region filter server to reserve or release the region. To enter the region, the region filter in an AMR requests a ticket with a topic---\texttt{/ticket} from the region filter server. The region filter returns the ticket with the topic---\texttt{/ticket} to the region filter server to leave the region. The region filter server responds to the region filter of each AMR based on  robot ID and region ID to identify the AMRs and regions. The region filter is implemented by   Algorithm~\ref{alg:regionfilter}.

\section{EXPERIMENTAL RESULTS}

\subsection{Hardware Settings}

\begin{figure}[t]
	\begin{center}
		\includegraphics[width=6cm]{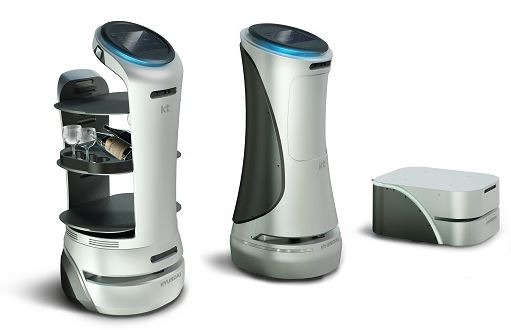}
		\caption{Hyundai Robotics AMRs lineup : Two 50-kg payload and one 100-kg payload delivery AMRs}
		\label{HyundaiRobotics}
	\end{center}
\end{figure}

For this study, the AMR (for testing) has been developed by Hyundai Robotics (Fig. \ref{HyundaiRobotics}). The AMR is equipped with a differential drive for wheel control, Wi-Fi for DDS, sensors for tracking, and potentially peripheral I/O for delivery. The differential drive is an all-in-one motor. The brushless motor, motor drive on the controller area network (CAN), wheel, and tire are included in the differential drive. The tracking sensor comprises a 2-D lidar scanner for positional sensing along with adaptive Monte Carlo localization (AMCL). The peripheral I/O can connect to the rotatable or prismatic type of the actuator. The goal is to use the AMR for transportation in factories or simple delivery services provided to customers in hotels. The robot software platform is ROS2 Eloquent, and DDS library is FastDDS\footnote{https://github.com/eProsima/Fast-DDS}. All experiments have been done with uncovered AMRs.

\subsection{Prohibition Filter}

\begin{figure}[t]
	\begin{center}
		\includegraphics[width=8.5cm]{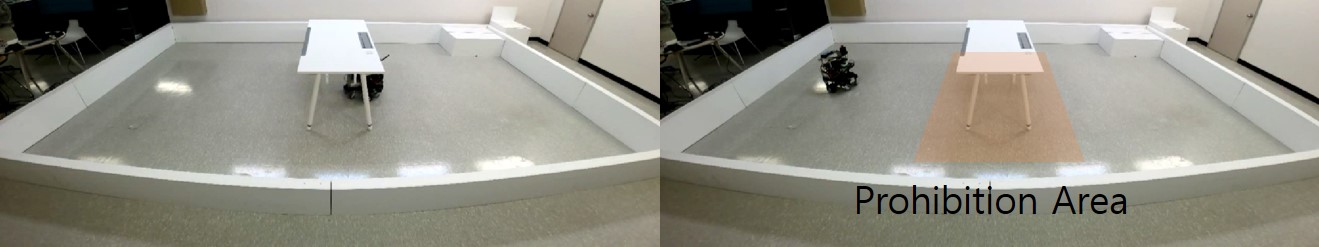}
		\includegraphics[width=8.5cm]{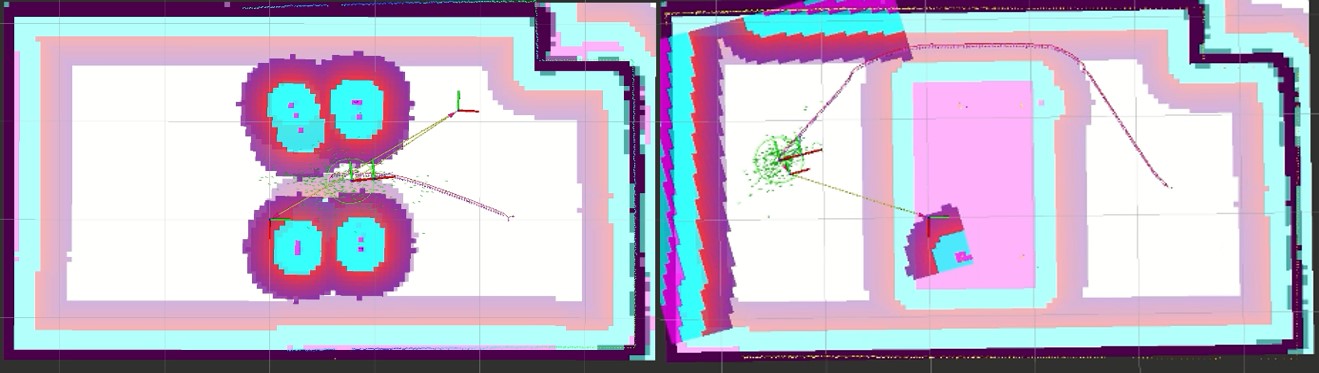}
		\caption{Comparison of the non-prohibition filters in the left and right sides}
		\label{prohibitionfilter}
		\vspace{-0.7cm}
	\end{center}
\end{figure}

As shown in Fig. \ref{prohibitionfilter}, a single AMR starts from the origin and reaches the destination point across the desk, which has four slender legs. The left part of the figure shows the AMR passing under the desk. As the cost map in the non-applied prohibition filter shows four groups of small dots of high cost values at the four legs, the AMR goes through  the  relatively low cost under the desk. In contrast, the prohibition filter prevents the AMR from the global or local planning under the desk. Keeping the desk area to preset a high cost value, the AMR moves  away from  the desk, even though only four legs are detected by the lidar sensor. 

\subsection{Lane Filter}

\begin{figure*}[b]
	\begin{center}
		\includegraphics[width=18cm]{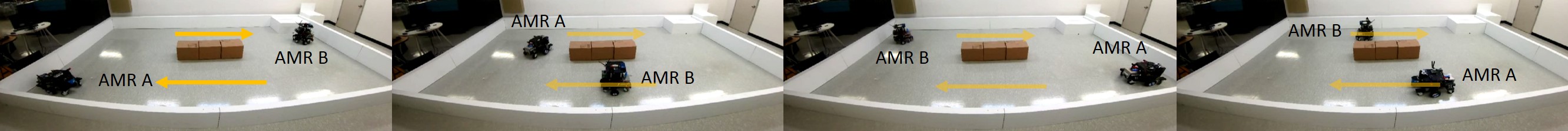}
		\caption{AMR A and B around three boxes under the lane filter.}
		\label{lanefilter}
		\vspace{-0.3cm}
	\end{center}
	
\end{figure*}

\begin{figure*}[b]
	\begin{center}
		\includegraphics[width=18.12cm]{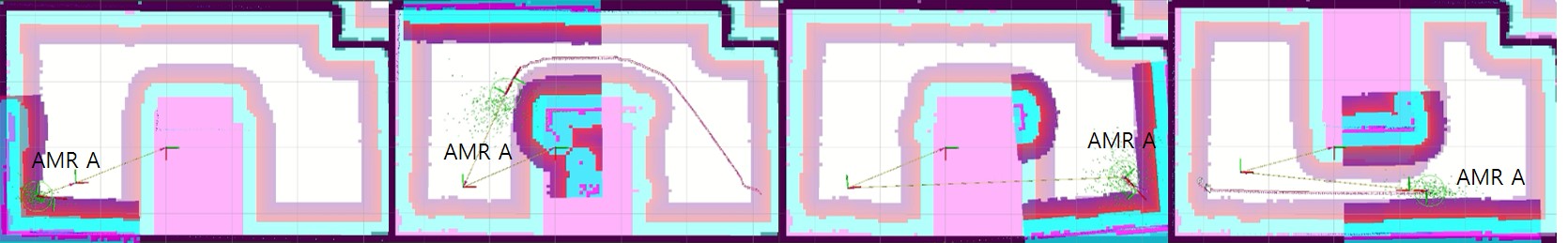}
		\includegraphics[width=18cm]{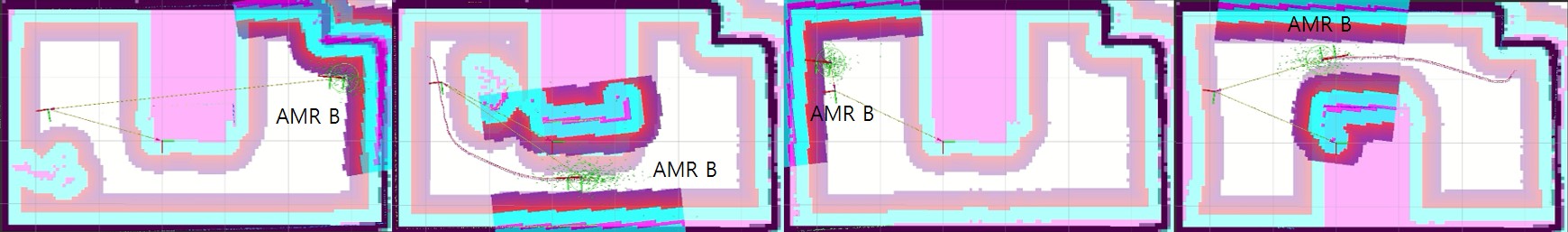}
		\caption{The cost maps of AMR A and B around three boxes under the lane filter.}
		\label{lanefiltercost}
		\vspace{-0.3cm}
	\end{center}
\end{figure*}

For testing  the lane filter, we have tried two AMRs to leave their starting points and reach the destination points. As shown in Fig. \ref{lanefilter},   three boxes are located together  at the center of the white rectangular area. There are two designated direction areas, which are divided by three boxes. Back  and front parts   of the three boxes are  set toward the right and lefts directions, respectively. The path between the starting and destination points is straight; hence, the AMRs can drive along the straight direction without the filter. However, the AMRs have turned around the three boxes and finally reached their destinations under  the lane filter. Based on the analysis of the cost map of each AMR shown in Fig. \ref{lanefiltercost}, it is found that lane areas depend on the cost maps. For AMR A, the cost of the front  lane increases as the direction cosine between AMR A and the lane directions is negative, since  AMR A faces the right direction. For AMR B, the opposite direction cosine is applied. Although the U-shaped navigation planning is inefficient with regard to energy or time-based planning, AMRs are forced to drive one way under  the lane filter.

\subsection{Fleet Layer}

\begin{figure*}[]
	\begin{center}
		\includegraphics[width=18cm]{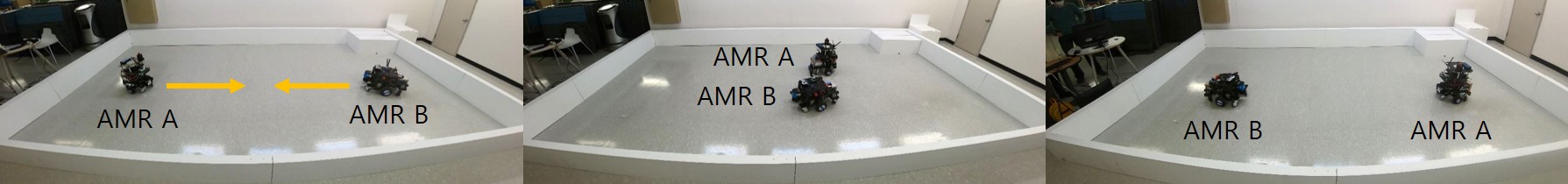}
		\caption{AMR A and B under the fleet layer.}
		\label{fleetlayer}
		\vspace{-0.3cm}
	\end{center}
\end{figure*}

\begin{figure*}[]
	\begin{center}
		\includegraphics[width=13.5cm]{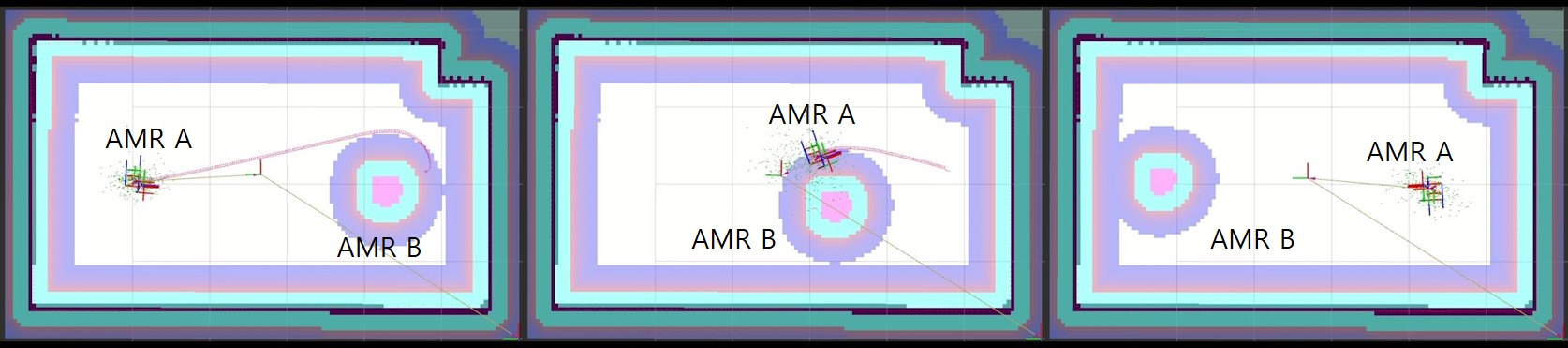}
		\includegraphics[width=13.5cm]{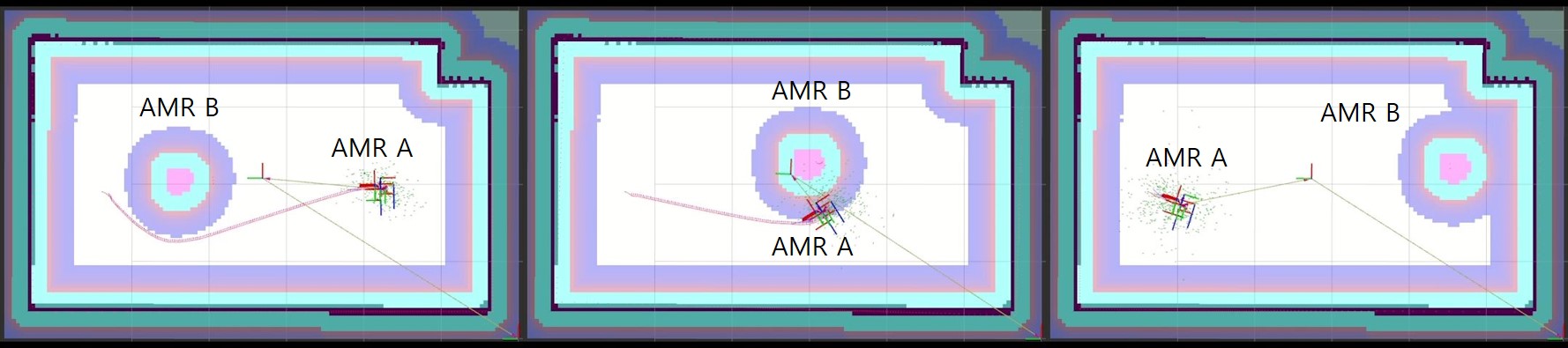}
		\caption{The cost maps of AMR  A and B under the fleet layer.}
		\label{fleetlayercost}
	\end{center}
\end{figure*}

As shown in Fig. \ref{fleetlayer}, two AMRs start from their origins to reach the origins of the opponents. To avoid collision, each AMR receives the position of the opponent AMR. Then, the fleet layer draws the opponent AMR based on the circle-shaped cost map (see Fig. \ref{fleetlayercost}). To navigate with low cost values, AMRs analyze global and local navigation planning while avoiding high-cost locations. As a result, the two AMRs avoid collision.

\begin{figure*}[]
	\begin{center}
		\includegraphics[width=13.5cm]{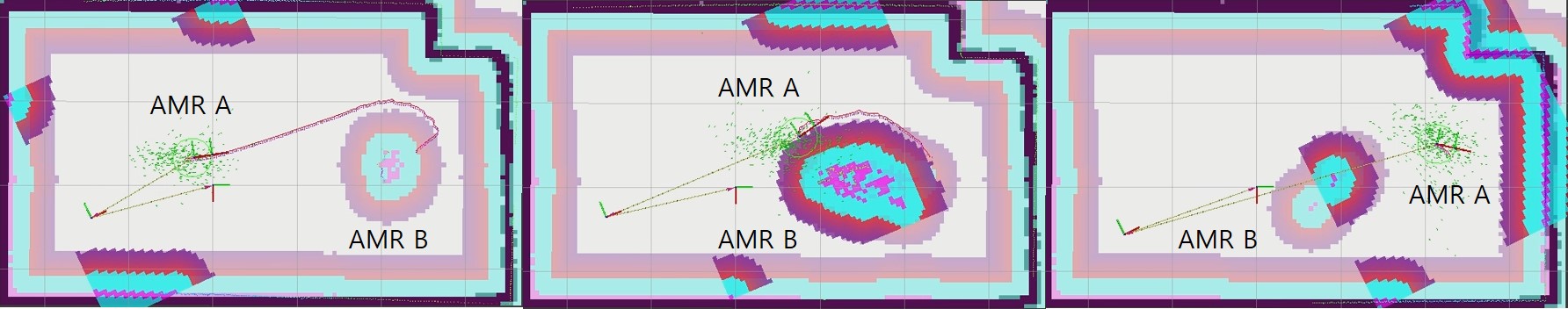}
		\includegraphics[width=13.5cm]{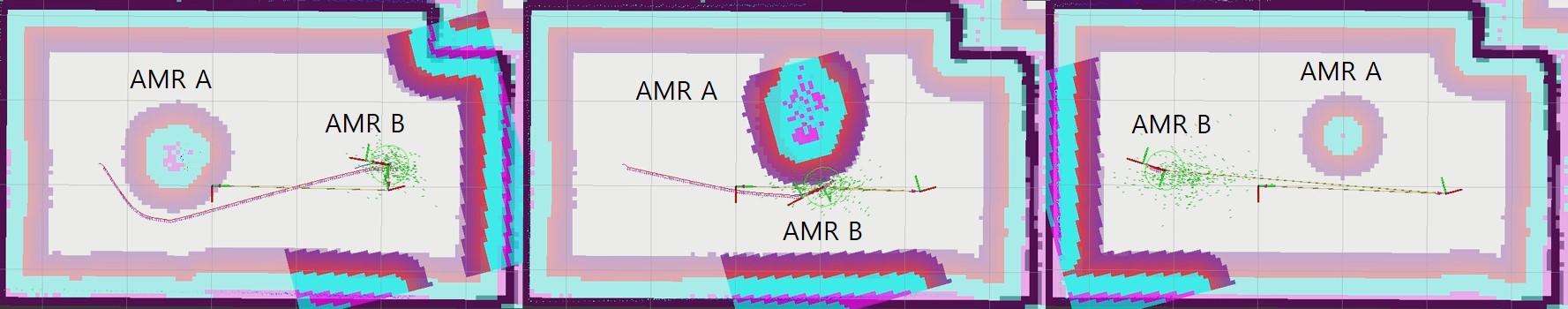}
		\caption{ The cost maps of AMR  A and B estimated by the lidar. }
		\label{obstaclelayercost}
	\end{center}
\end{figure*}

To compare the fleet layer to the AMR detection by the lidar, cost maps of AMR A and B estimated by the lidar are presented in Fig. \ref{obstaclelayercost}. First of all, it is found that the radius of the opponent AMR keeps changing. Moreover, we found that the opponent AMR is not moving anymore after turning against each other due to the limited lidar sensing angle as shown in the third part of Fig.~\ref{obstaclelayercost}. 

\begin{figure*}[]
	\begin{center}
		\includegraphics[width=18cm]{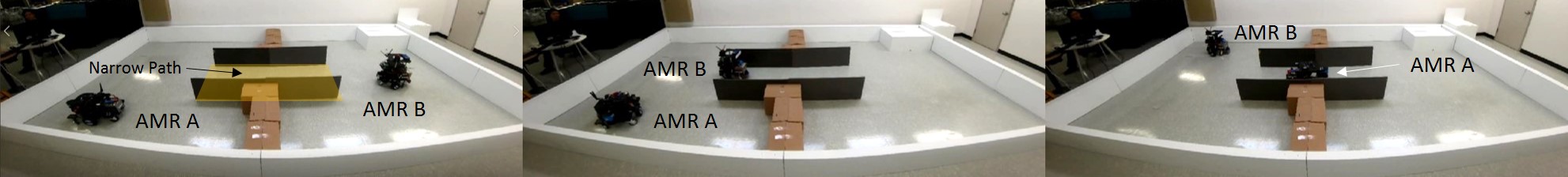}
		\caption{AMR  A and B  in the narrow path under the region filter.}
		\label{narrowpath}
		\vspace{-0.3cm}
	\end{center}
\end{figure*}

\begin{figure*}[]
	\begin{center}
		\includegraphics[width=13.5cm]{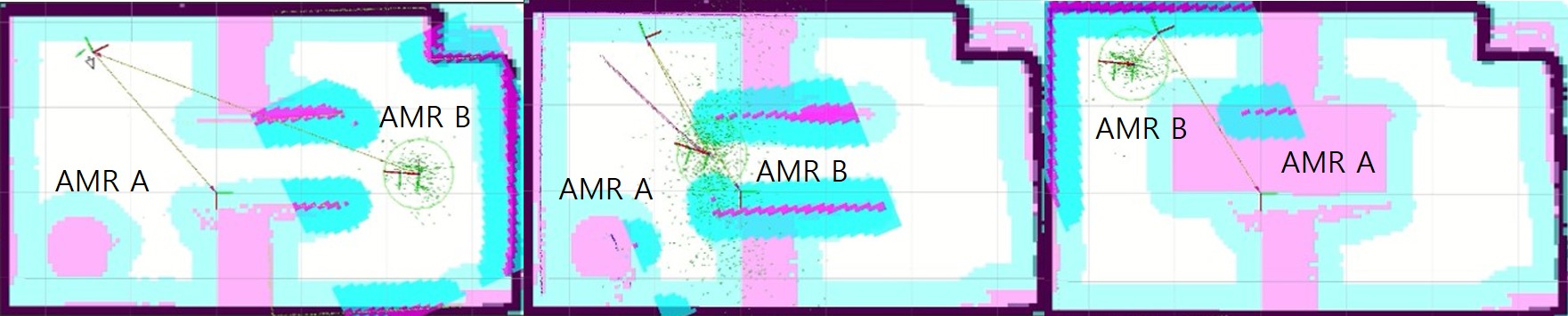}
		\includegraphics[width=13.5cm]{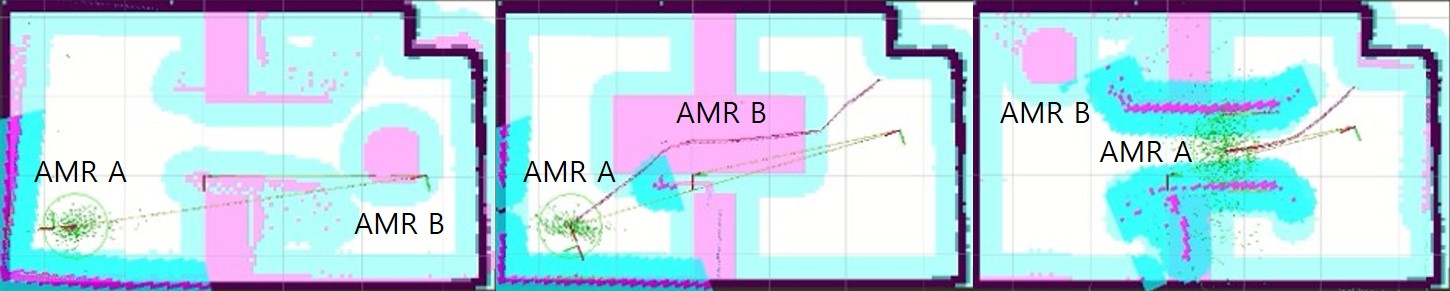}
		\caption{The cost maps of AMR B and A in the narrow path under the region filter.}
		\label{narrowpathcost}
	\end{center}
\end{figure*}
\begin{figure*}[]
	\begin{center}
		\includegraphics[width=18cm]{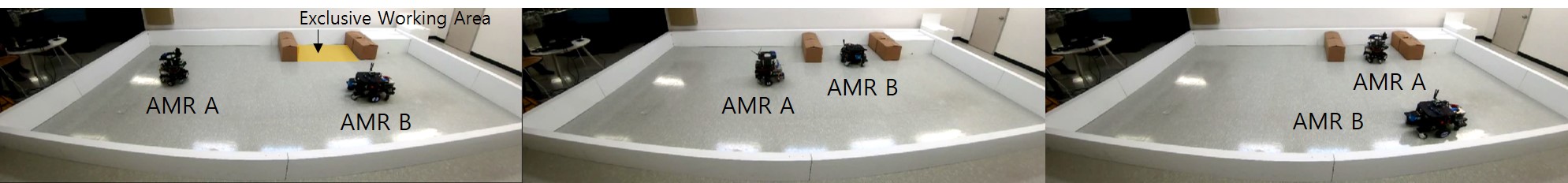}
		\caption{AMR  A and B    in the exclusive working region under the region filter.}
		\label{exclusiveworkingregion1}
	\end{center}
\end{figure*}

\begin{figure*}[]
	\begin{center}
		\includegraphics[width=13.5cm]{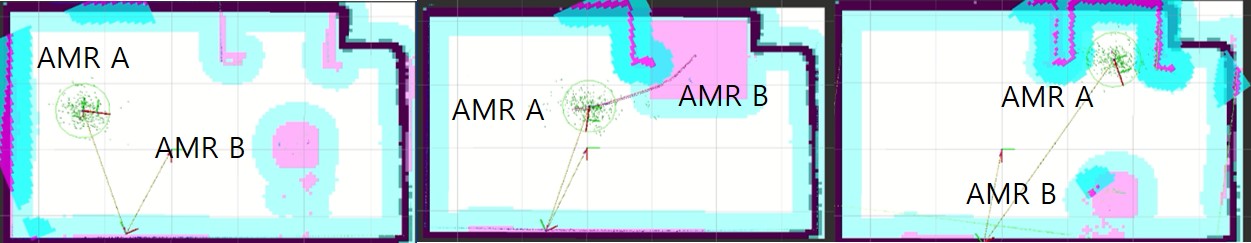}
		\includegraphics[width=13.5cm]{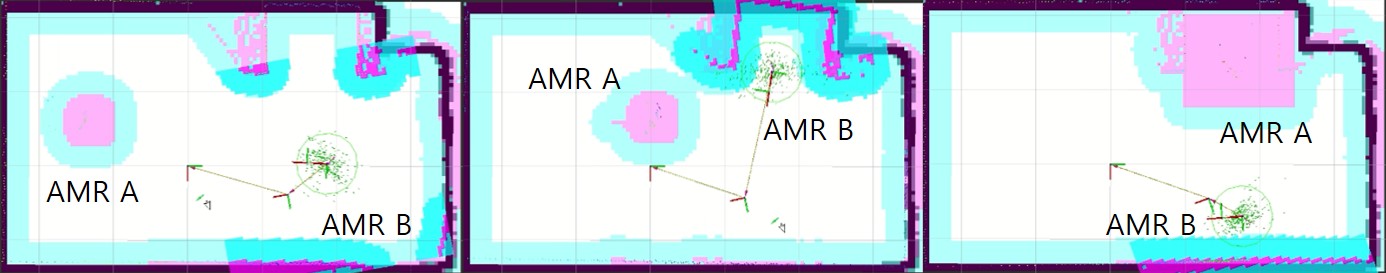}
		\caption{The cost maps of AMR  A and B in the exclusive working region under the region filter.}
		\label{exclusiveworkingregion}

	\end{center}
\end{figure*}

\subsection{Region Filter}

\subsubsection{Narrow path}
As shown in Fig. \ref{narrowpath}, multiple AMRs should pass through the narrow path whose width can only accommodate a single AMR. Two AMRs may inevitably collide with each other or stop infinitely,  with    deadlock. Once the destinations are set to the opposite locations  of two AMRs, they  shall perform global navigation planning. Providing higher priority with an AMR for passing through the path, the cost of the narrow path region can be differently set. Analyzing cost maps in Fig. \ref{narrowpathcost}, AMR  A and B try to drive the diagonally positioned goals through the narrow path. First, the narrow path opens for AMR B, which has higher priority. After AMR B arrives  to the  designated goal, AMR~A starts    passing  through the path when the narrow path lower the cost for  AMR A. 

\subsubsection{Exclusive working area}
We consider an AMR working in a designated area where other AMRs are prohibited. Considering this place as an exclusive working area, the AMR can use the place while reserving it to the region filter server. 
As shown in the first image of Fig. \ref{exclusiveworkingregion1}, two AMRs try to enter the exclusive working area, which is surrounded by six yellow boxes attached to the white wall. As shown in the second image,  AMR~B with higher priority enters the area, and   AMR~A waits around the U-shaped area. After finishing the task, AMR~B  leaves the area, and   AMR~B finally enters the area and performs the given task.
At the beginning, from AMR A's point of view, the exclusive working area is the prohibition area   (see Fig. \ref{exclusiveworkingregion}), since AMR~A has a lower priority. After AMR B leaves the area,  the region filter server releases the area from the prohibition area to a  free area for AMR~A. 

\section{CONCLUSION}
The traffic management system of AGVs cannot be used for AMRs    since   AMRs have no guided paths. 
Instead the cost map has been widely used in the navigation planning of AMRs. In our approach, for multiple AMRs, the layered  cost map data is proposed  to be shared among the AMRs to achieve traffic management. In particular, a DDS is adopted to share map  or reserve tickets. In this letter, we successfully demonstrated a single  AMR  tested with regard to the prohibition filter, and two AMRs   tested with regard to the one-way direction, narrow path, and exclusive working areas in real-world situations. To test the fleet layer, two AMRs were  tested for collision avoidance. 
The video of the experimental results is available at: \url{https://youtu.be/M--RUWZmbow/}.

\bibliographystyle{IEEEtran}
\bibliography{IEEEabrv,references}

\end{document}